\documentclass[runningheads,a4paper]{llncs2e/llncs}
\usepackage{amssymb}
\usepackage{graphicx}
\usepackage{subfigure}
\usepackage{multirow}

\usepackage{algorithm} 
\usepackage{algorithmic} 
\usepackage{booktabs}

\begin{document}

\mainmatter  

\title{A Graph Traversal Based Approach to Answer Non-Aggregation Questions Over DBpedia}

\titlerunning{Answering Non-Aggregation Questions Over DBPedia}

%
%
\author{Chenhao Zhu$^1$
\and Kan Ren$^1$
\and Xuan Liu$^1$ \and \\
    Haofen Wang$^2$
\and Yiding Tian$^1$
\and Yong Yu$^1$}
\authorrunning{C. Zhu et al.}

\institute{$^{1}$Shanghai Jiao Tong University, Shanghai, China\\
$^1$\{chzhu, kren, liuxuan0526, killa, yyu\}@apex.sjtu.edu.cn \\
$^{2}$East China University of Science and Technology, Shanghai, China\\
$^2$ whfcarter@ecust.edu.cn
}
%
%

\toctitle{A Graph Traversal-based Approach to Answer Non-Aggregation Questions Over DBpedia}
\tocauthor{C. Zhu et al.}
\maketitle

\begin{abstract}
We present a question answering system over DBpedia,
filling the gap between user information needs expressed in natural language and a structured query interface expressed in SPARQL over the underlying knowledge base (KB).
Given the KB, our goal is to comprehend a natural language query and provide corresponding accurate answers.
Focusing on solving the non-aggregation questions, in this paper, we construct a subgraph of the knowledge base from the detected entities and propose a graph traversal method to solve both the semantic item mapping problem and the disambiguation problem in a joint way.
Compared with existing work, we simplify the process of query intention understanding and pay more attention to the answer path ranking. We evaluate our method on a non-aggregation question dataset and further on a complete dataset. Experimental results show that our method achieves best performance compared with several state-of-the-art systems.

\end{abstract}

\section{Introduction}
\label{sec:introduction}
Nowadays great volume of linked data has been produced efficiently in both research and industrial areas,
such as DBpedia \cite{lehmann2014dbpedia}, YAGO \cite{suchanek2007yago}, Freebase \cite{bollacker2008freebase}, Google's Knowledge Graph and Microsoft's Satori.
Each of them contains a wealth of valuable knowledge stored in the form of predicate-argument structures, e.g., (\textit{subject}, \textit{predicate}, \textit{object}) triples.
Meanwhile, the quality of linked data (coverage and accuracy) is also increasing effectively with the help of well-designed research work and community efforts.
Consequently, a lot of work for various purposes have been developed by taking linked data as the underlying knowledge base.

However, surfing linked data requires ontological knowledge beforehand.
Even SPARQL is the most common query language of RDF data,
reading and surfing linked data web requires professional skills and extra learning cost, which makes common people unwilling to deeply browse.
It is crucial to propose techniques to fill the gap between users information needs and implicit data models including schema and instances.

Natural language Question Answering over Linked Data (QALD) may commendably achieve this goal while maintaining advantages of knowledge base.
Lopez et al. \cite{lopez2011question} surveyed the trend of question answering in semantic web and revealed some challenges as well as opportunities in natural language question answering.

Generally speaking, the main challenge of understanding a query intention in a structural form is to solve two problems, which are semantic item mapping and semantic item disambiguation. Semantic item mapping is recognizing the semantic relation topological structures in the natural language questions and then semantic item disambiguation is instantiating these structures regarding a given knowledge base. Unger et al. \cite{unger2012template} relies on parsing a question to formulate a SPARQL template to capture the intention of a user query. This template is then instantiated using statistical entity identification and predicate detection. He et al. \cite{he2014question} combines Markov networks with first-order logic in a probabilistic framework to achieve the goal of semantic mapping. Zou et al. \cite{zou2014natural} proposes a method to jointly solve semantic item mapping and disambiguation problems by reducing question answering to a subgraph matching problem.
However, they all focus on the semantic item mapping problem by adopting well-designed templates or complex model. In contrast, we simplify the semantic item mapping problem and pay more attention to solving the disambiguation problem.

In this paper, we propose a graph traversal-based method to solve semantic item mapping and disambiguation problems.
We solve the semantic item mapping problem in a simple way by parsing the question text to generate matched topological structures.
Based on these structures, we start from the detected entities in a question text.
Then we traverse from these entities to find connected predicates and resources in the knowledge base.
Next our approach uses a jointly ranking algorithm to solve the disambiguation problem.
Meanwhile we implement a constraint matching assessment of the answer type to find the best answer.

Since entities contain the most important semantic information in a natural language query,
the intuition is that we may get the right answer by finding the most suitable path in the knowledge base around the detected entities.

\begin{figure}[!htb] \centering{
\includegraphics[scale=0.4]{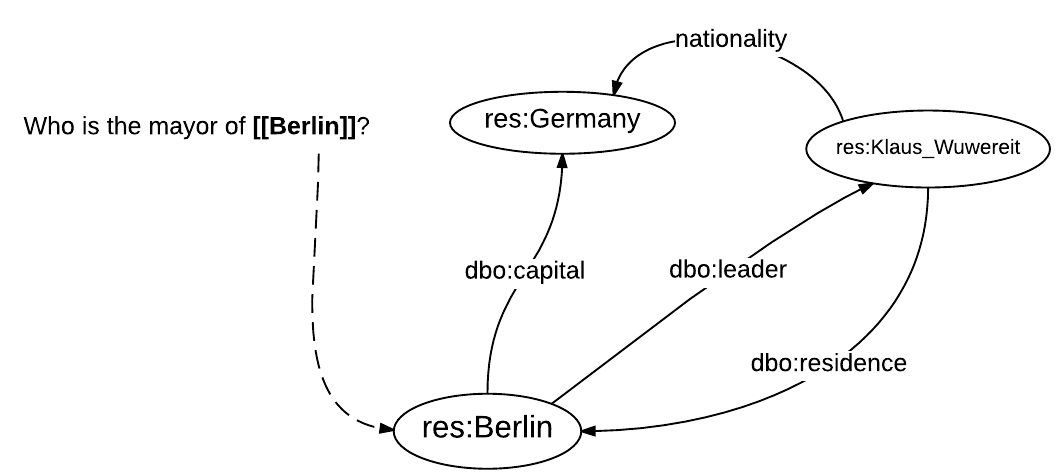}}
\caption{Graph Traversal Example}
\label{fig:subgraph-example}
\end{figure}

Take the question \emph{``Who is the mayor of Berlin?"} as an example.
As is shown in Figure \ref{fig:subgraph-example},
firstly we find the mention \textit{``Berlin"} and link it to the resource \textbf{res:Berlin} in the knowledge base.
Then our system traverses the subgraph around the resource and calculates ranking scores of the connected predicates.
In this case, we find that \textbf{dbo:leader} matches better since its label \textit{``leader"} has higher relatedness with \textit{``mayor"}.
Next our system makes a judgement that the traversing process stops here and \textbf{res:Klaus\_Wowereit} matches the answer type constraint.
So we return \textbf{res:Klaus\_Wowereit} as the final answer.

The contributions in this paper are summarized as follows:
\begin{itemize}
\item We present an approach for answering non-aggregation questions over DBpedia, which fills the gap between user information needs expressed in natural language and a structured query interface expressed in SPARQL.
\item We present an approach which is simple in structure and employs relatively lightweight machinery compared with existing work concentrating on complex models and training.
\item We compare our approach with several state-of-the-art systems on public dataset and achieve the best performance on non-aggregation questions. Moreover, we extend our approach to the complete dataset and also achieve the best performance.
\end{itemize}

The rest of this paper is organized as follows.
Section \ref{sec:relatedwork} briefly describes related work.
Section \ref{sec:framework} introduces the proposed graph traversal approach in detail.
Section \ref{sec:experiment} shows our experiments, including dataset collection, evaluation metric, comparison between our approach and several state-of-the-art work and error analysis.
Finally, Section \ref{sec:conclusion} concludes the paper and points out the future work.

\section{Related Work}
\label{sec:relatedwork}
Question answering (QA), which is to return the exact answers to a given natural language question, is a challenging task and has been advocated as a key problem for advancing web search. Previous work is mainly dominated by keyword based approaches, while recent blossom of large-scale knowledge bases have enable numerous KB-based systems. A KB-based QA system answers a question by directly querying structured knowledge knowledge, which can be retrieved using a structured query engine.

Recently many works have been published in this field.
Apart from \cite{he2014question}, \cite{unger2012template}, \cite{zou2014natural} discussed in Section \ref{sec:introduction},
PowerAqua \cite{lopez2012poweraqua} proposes a natural language user interface making people query and explore semantic web content more convenient.
Two research work \cite{yahya2012natural,yahya2013robust} present an ILP(Integer Linear Programming) method to translate a natural language question into a structured SPARQL query.
The Paralex system \cite{fader2013paraphrase} studies question answering as a machine learning problem and induces a function that maps open-domain questions to queries over a database of web extractions.
Meanwhile, Shekarpour et al. \cite{shekarpour2013question} presents an approach for question answering over a set of interlinked data sources. This approach firstly employs a Hidden Markov Model to determine the most suitable resources for a use-supplied query and secondly constructs a federated formal query using the disambiguated resources and linking structure of underlying datasets.
And two research work \cite{berant2013semantic,berant2014semantic} develop semantic parsing techniques that map natural language utterances into logical form queries, which can be executed on a knowledge base.
Xu et al. \cite{xu2015longest} develops a transition-based parsing model to do semantic parsing for aggregation questions.

However, most of these methods focus on translating a question to a SPARQL query. Meanwhile many methods need many well designed manual rules. Also lots of them focus on recognizing the inherent structure of user's query intention using different semantic parsing techniques implemented by complex models. In contrast, we aim to finding the most appropriate path rather than generate SPARQL query templates directly. And we simplify the process of query intention understanding and pay more attention to the answer path ranking. Our approach is simple in structure but effective in terms of performance.

\section{Framework}
\label{sec:framework}
\begin{figure}[!htb] \centering{
\includegraphics[scale=0.6]{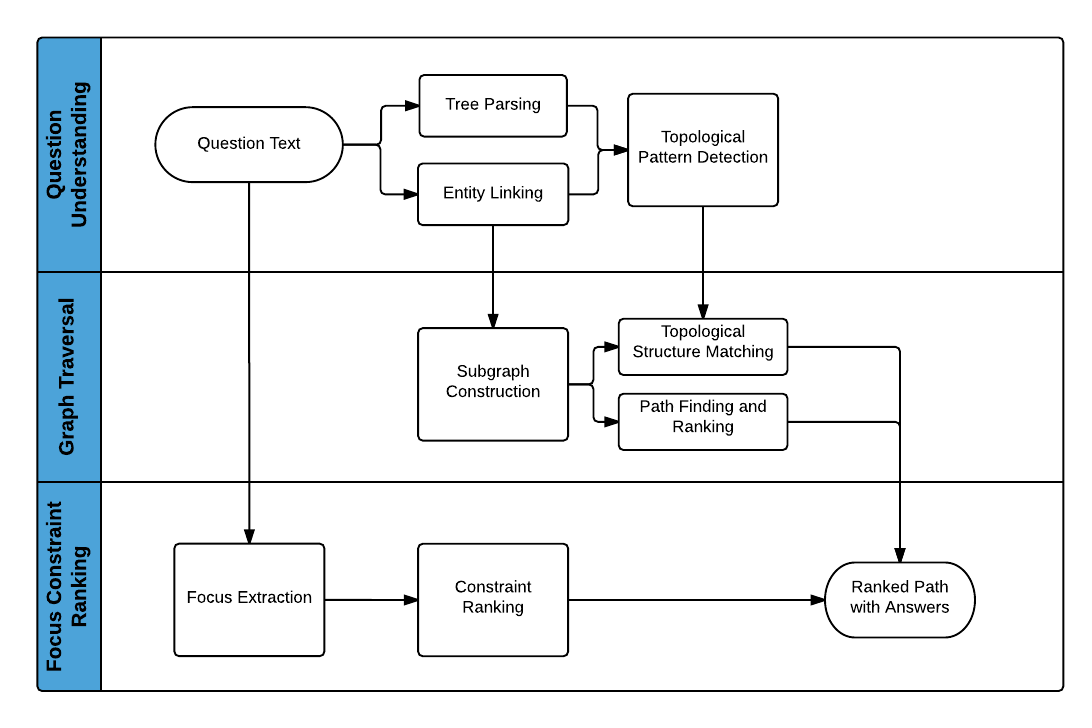}}
\caption{Overall Framework}
\label{fig:framework}
\end{figure}
The overall framework is shown in Figure \ref{fig:framework}.
Our approach aims to find the most appropriate path in the knowledge base rather than generate SPARQL query templates directly.
The whole process contains three phases:
\begin{itemize}
    \item \textbf{Question Understanding}: In this phase, the system detects the query's topological pattern, trying to capture its intention.
        To achieve this goal, we use an entity linking method to detect the mention-entity pairs, of which the mention is used for the phrase boundary identification.
        And next we build a list of topological patterns to discover the structure by taking advantage of the parsing result of the query.
    \item \textbf{Graph Traversal}: In this phase, we firstly build a subgraph of the underlying knowledge base rooted from entities we've found in last step.
        Then we use a jointly ranking method to find the most appropriate traversal path in the subgraph.
        The topological structure is used for semantic item mapping and judging traversal stop condition.
    \item \textbf{Focus Constraint}: We extract a phrase describing the answer directly from the query, which is called a \textit{focus}.
        Then we use this information to help modify final path ranking scores.
\end{itemize}

We solve the semantic item mapping problem during the question understanding phrase and  the disambiguation problem in the next two phrases. After above three phases, the overall path candidates ranking list is obtained. The answers found along the path with highest score will be returned.

\subsection{Question Understanding}
\begin{figure}[!htb] \centering{
\includegraphics[scale=0.6]{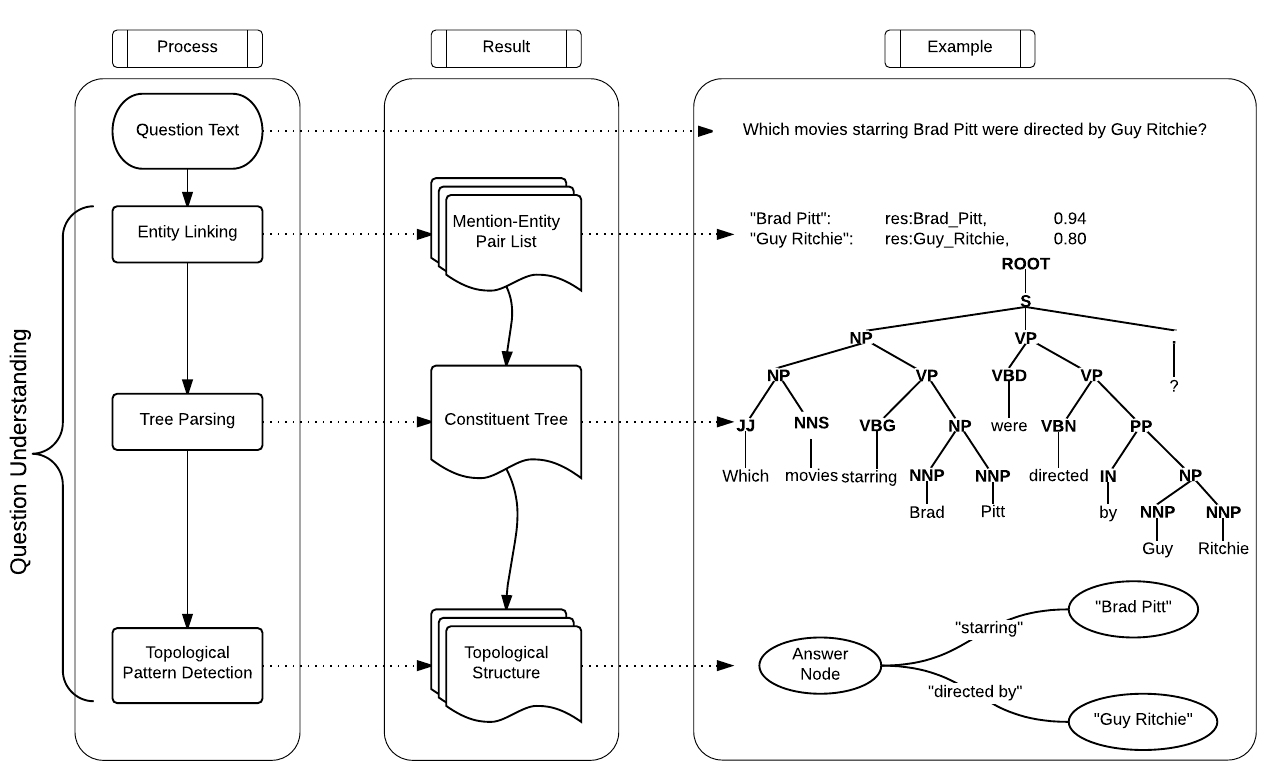}}
\caption{Process of Question Understanding with Question Example}
\label{fig:question-understanding}
\end{figure}

In this phase, we focus on question parsing and topological structure extraction.
Our system digests a natural language query and outputs its corresponding topological structure.
Figure \ref{fig:question-understanding} shows the whole process and results along with an example.

Given a question text, we firstly use an entity linking method to detect mentions in the query and link them to the resources in the knowledge base.
In this step we consider class entities and category entities such as \textbf{res:Actor} are of little importance. So in the entity linking phrase, we discard those corresponding mentions.
Meanwhile, each mention-entity pair will be attached with a confidence score.
We set a global threshold to discard those linking results with relatively low confidences.
As in the example, our system detects two mentions \textit{``Brad Pitt"} and \textit{``Guy Ritchie"} with corresponding entity results and confidence scores.
In our experiment, we use the Wikipedia Miner tool\footnote{http://wikipedia-miner.cms.waikato.ac.nz/} \cite{milne2013open} to detect the mentions from the question text and get the corresponding entity linking results. Empirically the parameter of min-Probability is set as 0.15.

In the next step, we extract topological structure of user intention with regard to our topological patterns.
We start from the constituent tree of a question text.
And table \ref{tab:pattern-list} lists our topological patterns.

Each pattern captures one form of relationship between two arguments.
For example, the pattern \textit{VB $\to$ VB+NP} means the VB and NP on the right side are the children of the VB on the left side with regard to the constituent tree.
We may derive the binary relation with arguments from it.
The third column in the table demonstrates the extraction result of the given example.
While ANSNODE is a wildcard representing the current answer we are looking for.

\begin{table}\centering{
\caption{Topological Pattern List}
\label{tab:pattern-list}
    \begin{tabular}{ | l | p{3cm} | p{4cm} | p{5cm} |}
    \hline
    ID & Pattern & Example & Extraction Result \\ \hline
    1 & VB $\to$ VB+NP & Who produces Orangina? & ANSNODE - ``produces" - ``Orangina" \\ \hline
    2 & VP $\to$ VB+PP & Which television shows were created by John Cleese? & ANSNODE - ``created by" - ``John Cleese" \\ \hline
    3 & NP $\to$ NP+PP & Who is the mayor of Berlin? & ANSNODE - ``mayor of" - ``Berlin" \\ \hline
    4 & SQ $\to$ VB+NP+VP & When was Alberta admitted as province? & ANSNODE - ``admitted as province" - ``Alberta" \\
    \hline
    \end{tabular}
    }
\end{table}

In our running example, our system extracts two relationships using Pattern 1 and 2.
The extraction results are presented in Figure \ref{fig:question-understanding}.

We use a recursive method to discover all the relations in the question text.
Note that there is one case we should handle carefully.
If the entity linking phase produces a mention that is fit for one topological pattern, our extraction stops the recursive process immediately.
Take question \textit{``Who wrote the book The Pillars of the Earth?"} as an example.
Pattern 3 is matched in the phrase \textit{``The Pillars of the Earth"} while it is a piece of a mention detected by the entity linking method.
So our recursive algorithm will skip processing it.
In our experiment, we use the Stanford Parser\footnote{http://nlp.stanford.edu/software/index.shtml} \cite{manning2014stanford} to parse the question text and generate the corresponding constituent tree.

\subsection{Graph Traversal}
\begin{figure}[!htb] \centering{
\includegraphics[scale=0.5]{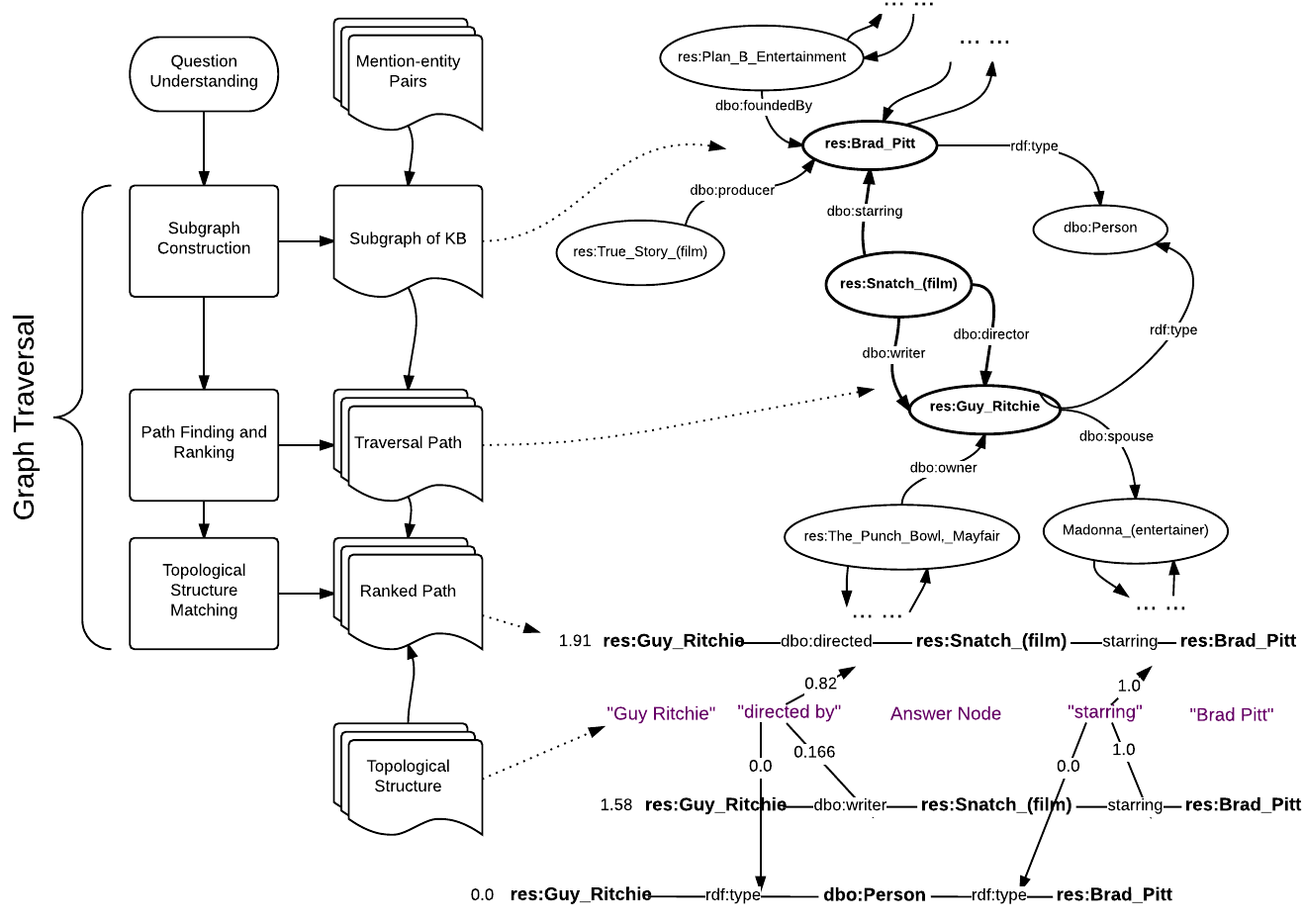}}
\caption{Process of Graph Traversal with Question Example}
\label{fig:graph-traversal}
\end{figure}

Our approach does not generate any SPARQL templates for a given question text.
We just traverse the knowledge base from entities found in the entity linking phase.
The main three steps are subgraph construction, path finding and topological structure matching.
The whole process has been presented in Figure \ref{fig:graph-traversal}.

In the first step, we begin from linked entities in the knowledge base and construct a subgraph surrounding with them.
Given an entity $e$ in the mention-entity list, we root the graph from $e$ and expand one layer.
By this means we may obtain all the entities including resources and classes around $e$ within 1 step distance.
Then we retrieve more entities around entity $e$ with 2 step distance.
In the end, we have $K$ layers subgraphs in the knowledge base around entity $e$.
Here $K$ is the longest distance between two nodes in topological structure.

In our running example, the entity list provides two resources \textbf{res:Brad\_Pitt} and \textbf{res:Guy\_Ritchie}.
We begin from each of them and construct a subgraph on the right part in Figure \ref{fig:graph-traversal}.
Here $K$ is 1. And the two subgraphs of the linked entities here are combined into a single one.

After the subgraph construction, we implement a path finding and ranking algorithm to get the answer.
The method also begins from detected entities.
The goal is to find the path which is the most appropriate to match the information in the question text.

We start from a known entity $e$ and make one step ahead.
After that we obtain many one step pathes rooted at $e$.
Our system calculates the semantic similarity between the predicates around the entity $e$ and the phrase text from the edge in the topological structure.
Then we will make one more step outward and get many two-length pathes.
For these new predicates, we repeat the same calculation of semantic similarity between predicate labels and the corresponding phrase text.
To the end we may obtain many pathes starting from entity $e$ with ranking scores of each edge.

One question is how we judge that we will stop expanding outward. Here we make two stop conditions.
The first one is simple that we will stop finding when it gets to the outmost layer of the subgraph.
The second condition is based on the topological structure.
Our system will decide that if the path obtained above matches the topological structure, and discard those not matching ones.

As is shown in Figure \ref{fig:question-understanding}, the topological structure is triangular.
Two known entities link to the answer node with two pieces of phrase text.
The answer node is a wildcard representing what we search for in the knowledge base.
We use the phrase text of the edge in topological structure to match a predicate.

In Figure \ref{fig:graph-traversal}, our approach finds three different pathes.
Each path contains two known entities.
Our ranking module calculates scores for each predicate with phrase texts.
At last our system will jointly rank each path with regard to ranking scores for predicates.
The ranking list shows that the most appropriate path is the first one and \textbf{res:Snatch\_(film)} is our answer (till now).

\subsection{Focus Constraint}
As is defined in section \ref{sec:framework}, a \textit{focus} is a phrase in the question text describing the answer directly.
For example, \textit{``television shows''} is the focus of \textit{``Which television shows were created by John Cleese?''}.
It implies that the answer is a type of television show.
In this part, we extract the focus from a question text for the calculation of additional ranking scores to pathes obtained above.

Intuitively, we may derive focus information from interrogatives.
More specifically, we extract \textit{person} and \textit{organization} from \textit{``who''}, \textit{place} from \textit{``where''} and \textit{date} from \textit{``when''}.

Besides, we extract a focus based on the POS tags of a question.
In our approach, the longest noun phrase after the interrogative part of a question will be considered as the focus.
The interrogative part means the phrase in a query used to start the question.
Here the interrogative part could be \textit{``Give me all''}, \textit{``What''}, \textit{``Which''} and so on.
We extract the first word after the interrogative part with the POS tag ``NN'' (or other NN-like tags) to the last word having continuous ``NN'' tag as the focus.
In our running example, we extract \textit{``movies''} as the focus.

In the next step, we use the focus phrase to modify the predicate ranking result and calculate a matching score of the answer type.

For predicate ranking, we use the focus phrase as additional information to rank the last predicate.
The last predicate is the nearest predicate in the path to get to the answer entities.
For answer type constraint, we use the headword of the focus phrase and calculate the similarity between the headword and the type information of the answers.
Answer entities are obtained by ranked path above.
The similarity matching score will be added to the ranking score of the path.
Detail is discussed in section \ref{sec:path_ranking}.

Now that we've calculated the ranking score for each path with answer type similarity score, we may obtain the best answer of the question text.

\subsection{Path Ranking}
\label{sec:path_ranking}
A path score is composed of two parts, including a predicate score and a type score. Moreover, the predicate score is related to all predicates matched in the topological structure.
We can formulate a path score as Equation \ref{equ:path-score}.

\begin{equation}
\label{equ:path-score}
  PathScore ={\frac{1}{m}}\sum_{i=1}^m{PredicateScore_i} + TypeScore
\end{equation}

Next in this section, we will discuss the details about the predicate ranking and jointly path score calculation.

For predicate ranking, we consider the semantic similarity between a predicate and the phrase text extracted
from the topological structure. More specially, to the predicates leading to the final answers we add some information
from focus constraint since it also contributes to the predicate identification. The detail is showed in Algorithm \ref{alg:predicate_ranking}.
We use the UMBC Semantic Similarity Service\footnote{http://swoogle.umbc.edu/SimService/} \cite{han2013umbc} to calculate the semantic similarity between two words.
\begin{algorithm}[!htb]
  \caption{Predicate Ranking}
  \label{alg:predicate_ranking}
  \begin{algorithmic}[1]
  \REQUIRE
    A predicate $p$;
    Phrase text $t$;
  \ENSURE The Predicate $p$'s ranking score $s$;

  \STATE Find all labels $Labels$ of $p$ in RDF Repository;
  \FOR{each label $l\in Labels$}
    \FOR{each word $w$ of $l$}
        \FOR{each word $tw$ of $t$}
            \STATE Calculate the word semantic similarity $wss$ between $w$ and $tw$;
        \ENDFOR
        \STATE Set the maximal $wss$ as $w$'s score $ws$;
    \ENDFOR
    \STATE Set $l$'s score $ls$ as the arithmetic mean of each word's $wss$;
  \ENDFOR
  \STATE Set $s$ as the maximal $ls$;
  \end{algorithmic}
\end{algorithm}

For path ranking, we combine the information from the possible predicates, the topological pattern and the answer type.
If the question satisfies the one step path, the path ranking score is simply the sum
of the predicate ranking score and the answer type score generating from the focus constraint.
While, for the two step path ranking, it needs more work. The detail is show in Algorithm \ref{alg:path_ranking}.
Since the knowledge base is in a large scale and an entity has too many predicates, we need to restrict the candidate size for efficiency. In our experiment the maximal size of candidate predicates is set to 5 in each step during the phrase of path ranking.
\begin{algorithm}[!htb]
  \caption{Two Step Path Ranking}
  \label{alg:path_ranking}
  \begin{algorithmic}[1]
  \REQUIRE
    Predicate pairs $pps$ matching the corresponding topological pattern;
    Phrase Texts $pts$ of the corresponding predicates;
    Minimal semantic similarity threshold $\tau$ for each predicate;
  \ENSURE
    Path ranking scores;
  \FOR{each predicate pair $pp \in pps$}
    \STATE Calculate the two predicates' ranking score $s1$, $s2$ using $pts$;
    \IF{$s1 < \tau || s2 < \tau$}
        \STATE Remove this predicate pair;
        \STATE $\textbf{Continue}$;
    \ENDIF
    \STATE Set $pp$'s predicate pair Score $ps$ as the arithmetic mean of $s1$ and $s2$;
    \STATE
    \STATE Find all answers according to the path constructed by $pp$
    and calculate the type constraint score $ts$ according to their matching degree;
    \STATE
    \STATE Set the path's ranking score as the sum of $ps$ and $ts$;
  \ENDFOR
  \end{algorithmic}
\end{algorithm}

\section{Experiment}
\label{sec:experiment}
\subsection{Dataset}
QALD task \cite{cimiano2013multilingual} is the only benchmark for the KB-based QA problem. It includes both the underling knowledge base and the natural language questions. QALD is based on DBpedia knowledge base.

We use the QALD-3 test dataset\footnote{http://greententacle.techfak.uni-bielefeld.de/~cunger/qald/3/data/dbpedia-test.xml} in our experiments. This dataset has 99 questions in total. And it includes various types of questions. Table \ref{tab:question-type} lists the details. To build a non-aggregation questions dataset (QALD-3-NA), we filter the questions which need operations such as count, filter or order by, meanwhile, we filter the questions which have no answers in DBpedia or can not be solved using DBpedia singly. Then our new dataset QALD-3-NA has 61 questions in total.

\begin{table}\centering{
\caption{Question Type in QALD-3 test dataset}
\label{tab:question-type}
    \setlength{\tabcolsep}{3pt}
    \begin{tabular}{*{9}{c}}
    \hline
        Type & Non-aggregation & Count & Filter & Order By & Boolean & Out Of Scope\\ \hline
        Num & 61 & 4 & 7 & 4 & 7 & 16\\
        \hline
    \end{tabular}
    }

\end{table}

Considering the public dataset QALD-3 used by other state-of-the-art systems also contains aggregation questions while our approach focuses on answering non-aggregation questions, meanwhile few of state-of-the-art systems are publicly avaiable, we design two aspects of experiments to verify the validity of our method.

Firstly, we collect a dataset by selecting all non-aggregation questions from the public dataset. Then we compare our method with one state-of-the-art system gAnswer \cite{zou2014natural} on the new dataset, since gAnswer \cite{zou2014natural} also uses DBpedia as the underlying KB and offers an publicly online demo\footnote{http://59.108.48.18:8080/gAnswer/ganswer.jsp}, which makes it possible to serve as a contrast. This experiment is used to indicate our performance on non-aggregation questions directly.

Secondly, we make an experiment on the complete public dataset and compare our method with two state-of-the-art systems DEANNA \cite{yahya2012natural}, gAnswer \cite{zou2014natural} and all participating systems in the QALD-3 competition. If we achieve better performance, it can verify the validity of our method on non-aggregation questions from another point of view.

\subsection{Evaluation Metrics}
To enable the comparison with other state-of-the-art systems and the systems in QALD-3 competition, we adopt the same evaluation metrics used in QALD-3. That is to say, firstly, for each of the questions, we evaluate its precision, recall and F1-measure. Next we compute the overall precision and recall taking the average mean of all single precision and recall values, as well as the overall F1-measure \cite{cimiano2013multilingual}.


\subsection{Evaluation Results}
On the QALD-3-NA dataset, we compare our method with one state-of-the-art system gAnswer \cite{zou2014natural}. Table \ref{tab:evaluation-result-na} shows the evaluation result of average precision, average recall and average F-1 score. Meanwhile, it shows the number of question our system can answer, the number of right and partially right answers among them. We report the 30 questions which we can answer correctly in Table \ref{tab:correct-questions-total} and the 13 questions which we can answer partially in Table \ref{tab:correct-questions-partially}.

\begin{table}\centering{
\caption{Evaluation result on QALD-3-NA test dataset}
\label{tab:evaluation-result-na}
    \setlength{\tabcolsep}{3pt}
    \begin{tabular}{*{9}{c}}
    \hline
        & Total & Processed & Right & Partial & Avg.Recall & Avg.Precision & Avg.F-1 \\ \hline
        gAnswer demo & 61 & 38 & 21 & 7 & 0.41 & 0.45 & 0.42\\
        \textbf{Ours} & 61 & 53 & 30 & 13 & \textbf{0.67} & \textbf{0.61} & \textbf{0.61}\\
        \hline
    \end{tabular}
    }

\end{table}

To further indicate our performance, we apply our approach on the QALD-3 dataset compared with two state-of-the-art systems, namely, gAnswer \cite{zou2014natural} and DEANNA \cite{yahya2012natural}. Also, we compare our approach with all the participating systems in QALD-3 competition, whose results are reported in the QALD-3 overview paper \cite{cimiano2013multilingual}. Table \ref{tab:evaluation-result} shows the results.

\begin{table}\centering{
\caption{Evaluation result on QALD-3 test dataset}
\label{tab:evaluation-result}
    \setlength{\tabcolsep}{3pt}
    \begin{tabular}{*{9}{c}}
    \hline
        & Total & Processed & Right & Partial & Avg.Recall & Avg.Precision & Avg.F-1 \\ \hline
        \textbf{Ours (NA)} & 99 & 53 & 30 & 13 & 0.42 & 0.38 & 0.38\\
        \textbf{Ours (Total)} & 99 & 60 & 31 & 17 & \textbf{0.46} & \textbf{0.40} & \textbf{0.40}\\
        gAnswer demo & 99 & 50 & 23 & 11 & 0.30 & 0.30 & 0.28\\
        gAnswer \cite{zou2014natural} & 99 & 76 & 32 & 11 & 0.40 & \textbf{0.40} & \textbf{0.40}\\
        DEANNA \cite{yahya2012natural} & 99 & 27 & 21 & 0 & 0.21 & 0.21 & 0.21\\
        CASIA \cite{he2013casia} & 99 & 52 & 29 & 8 & 0.36 & 0.35 & 0.36\\
        Scalewelis \cite{joris2013scalewelis} & 99 & 70 & 32 & 1 & 0.33 & 0.33 & 0.33\\
        RTV \cite{giannone2013hmm} & 99 & 55 & 30 & 4 & 0.34 & 0.32 & 0.33\\
        Intui2 \cite{dima2013intui2} & 99 & 99 & 28 & 4 & 0.32 & 0.32 & 0.32\\
        SWIP \cite{pradel2013swip} & 99 & 21 & 15 & 2 & 0.16 & 0.17 & 0.17\\
        \hline
    \end{tabular}
    }
\end{table}

From Table \ref{tab:evaluation-result-na}, we can see that our approach achieves much better performance on the non-aggregation questions. Meanwhile, the result in Table \ref{tab:evaluation-result} also verifies the validity of our approach from another point of view. On the QALD-3 dataset, we show our two results. \textbf{Ours (NA)} is evaluated by adapting our performance on QALD-3-NA to the complete QALD-3 dataset, which simply multiplies the ratio of the question numbers of two dataset according to the evaluation metrics. That is to say, we set the precision, recall and F1-score as 0 on the other types questions and then get a global result. Although it is a little unfair to us, \textbf{Ours (NA)} outperforms most of state-of-the-art systems and only has a narrow gap to the best one of state-of-the-art systems. However,  \textbf{Ours (Total)} achieves best performance, especially in the evaluation of recall.

The reason why our F-1 scores on two datasets are not equal to the ratio of the question numbers of two dataset has two explanations. Firstly, for those questions which need count-operations, using our approach, we can get the results and what we need to do further is to simply count the number. Secondly, for those questions which need filter-operations, our approach can get right answers together with wrong answers which should  be filtered. The details of the contribution from different types of questions are showed in Table \ref{tab:contribution-analysis}.

\begin{table}\centering{
\caption{Contribution of different types of questions on QALD-3 dataset}
\label{tab:contribution-analysis}
    \setlength{\tabcolsep}{2pt}
    \begin{minipage}{\textwidth}
    \begin{tabular}{*{9}{c}}
    \hline
        Type & Total & Processed & Right & Partial & Avg.Recall & Avg.Precision & Avg.F-1 \\ \hline
        Non-aggregation & 61 & 53 & 30 & 13 & 0.42 & 0.38 & 0.38\\
        Count & 4 & 3 & 1 & 0 & 0.01 & 0.01 & 0.01\\
        Filter & 7 & 4 & 0 & 4 & 0.03 & 0.01 & 0.02\\
        Order By & 4 & 0 & 0 & 0 & 0 & 0 & 0\\
        Boolean & 7 & 0 & 0 & 0 & 0 & 0 & 0\\
        Out Of Scope & 16 & 0 & 0 & 0 & 0 & 0 & 0\\
        \textbf{Sum} & 99 & 60 & 31 & 17 & 0.46 & 0.40 & 0.41\footnote{Compared to 0.40 in Table\ref{tab:evaluation-result}, the inconsistency here is the result of a round-off error.}\\

        \hline
    \end{tabular}
    \end{minipage}
    }
\end{table}

\begin{table}\centering{
\caption{The QALD-3-NA Questions that can be answered correctly in our system}
\label{tab:correct-questions-total}
    \setlength{\tabcolsep}{6pt}
    \begin{tabular}{*{2}{l}}
    \hline
        ID & Questions\\ \hline
        Q3 & Who is the mayor of Berlin?\\
        Q4 & How many students does the Free University in Amsterdam have?\\
        Q7 & When was Alberta admitted as province?\\
        Q19 & Give me all people that were born in Vienna and died in Berlin.\\
        Q20 & How tall is Michael Jordan?\\
        Q22 & Who is the governor of Wyoming?\\
        Q24 & Who was the father of Queen Elizabeth II?\\
        Q30 & What is the birth name of Angela Merkel?\\
        Q35 & Who developed Minecraft?\\
        Q38 & How many inhabitants does Maribor have?\\
        Q42 & Who is the husband of Amanda Palmer?\\
        Q43 & Give me all breeds of the German Shepherd dog.\\
        Q44 & Which cities does the Weser flow through?\\
        Q45 & Which countries are connected by the Rhine?\\
        Q53 & What is the ruling party in Lisbon?\\
        Q54 & What are the nicknames of San Francisco?\\
        Q56 & When were the Hells Angels founded?\\
        Q58 & What is the time zone of Salt Lake City?\\
        Q65	& Which instruments did John Lennon play?\\
        Q66	& Which ships were called after Benjamin Franklin?\\
        Q68	& How many employees does Google have?\\
        Q71	& When was the Statue of Liberty built?\\
        Q74	& When did Michael Jackson die?\\
        Q76	& List the children of Margaret Thatcher.\\
        Q81	& Which books by Kerouac were published by Viking Press?\\
        Q83	& How high is the Mount Everest?\\
        Q85	& How many people live in the capital of Australia?\\
        Q86	& What is the largest city in Australia?\\
        Q98	& Which country does the creator of Miffy come from?\\
        Q100	& Who produces Orangina?\\
        \hline
    \end{tabular}
    }

\end{table}

\begin{table}\centering{
\caption{The QALD-3-NA Questions that can be answered partially in our system}
\label{tab:correct-questions-partially}
    \setlength{\tabcolsep}{6pt}
    \begin{tabular}{*{2}{l}}
    \hline
        ID & Questions\\ \hline
        Q2	& Who was the successor of John F. Kennedy?\\
        Q8	& To which countries does the Himalayan mountain system extend?\\
        Q17	& Give me all cars that are produced in Germany.\\
        Q21	& What is the capital of Canada?\\
        Q28	& Give me all movies directed by Francis Ford Coppola.\\
        Q29	& Give me all actors starring in movies directed by and starring William Shatner.\\
        Q41	& Who founded Intel?\\
        Q48	& In which UK city are the headquarters of the MI6?\\
        Q64	& Give me all launch pads operated by NASA.\\
        Q67	& Who are the parents of the wife of Juan Carlos I?\\
        Q72	& In which U.S. state is Fort Knox located?\\
        Q84	& Who created the comic Captain America?\\
        Q89	& In which city was the former Dutch queen Juliana buried?\\
        \hline
    \end{tabular}
    }

\end{table}

\subsection{Error Analysis}
Here we provide the error analysis of our approach. There are four key reasons for the error of some questions in our approach.
The first one is the entity linking error. In some cases, we fail to find the correct entities in a question text.
The second one is the semantic item mapping error.
It contains two aspects of reasons.
In some cases we fail to extract the structure correctly, and in other cases, since our current algorithm of finding predicates around an entity does not consider the subject or object role of entities detected at the beginning or generated as the intermediate result, we make some mistakes. For example,  in the case of \textit{``Who are the parents of the wife of Juan Carlos I?"}, we not only correctly get the parents of Juan Carlos I's wife, but also make a mistake by getting her children at the same time. This is the key reason that we have a relatively high recall compared with the precision. In this case, the recall equals 1 while the precision has a loss and only gives the value 0.40.
The third one is the path ranking error. We fail to rank the right path in the first place in some cases.
The fourth one we call it the restriction error. In our method we use the focus in a query to grade predicates and answer type constricts. However, a focus should be transferred to part of the final SPARQL query which restricts the answer type.
The percentage of each reason is showed in Table \ref{tab:error-reason}.

\begin{table}\centering{
\caption{Error Analysis}
\label{tab:error-reason}
    \setlength{\tabcolsep}{2pt}
    \begin{tabular}{*{9}{c}}
    \hline
        Type & Percentage\\ \hline
        Entity linking error & 26\%\\
        Structure extraction error & 16\%\\
        Semantic role error & 13\%\\
        Path ranking error & 13\%\\
        Restriction error & 29\%\\
        Others & 3\%\\
        \hline
    \end{tabular}
    }
\end{table}

\section{Conclusion and Future work}
\label{sec:conclusion}
In this paper, we propose a graph traversal-based approach to answer non-aggregation natural language questions over linked data. Our system starts from the detected entities and puts more attention on ranking the predicate paths. By translating the natural language question to a topological structure and mapping the structure to the linked data utilizing both the semantic features of the phrase similarity and type constraints. Compared with existing work, our method employs relatively lightweight machinery but has good performance.
In the future, we will adapt our method to answer aggregation questions, meanwhile we will try to answer a question by combining multiple knowledge bases to make our system more adaptable and more powerful.
\\\\ \textbf{Acknowledgments.} This work was partially supported by the National Science Foundation of China
(project No: 61402173) and the Fundamental Research Funds for the Central Universities
(Grant No: 22A201514045).

\bibliographystyle{llncs2e/splncs03}
\bibliography{paper}

\begin{thebibliography}{10}
\providecommand{\url}[1]{\texttt{#1}}
\providecommand{\urlprefix}{URL }

\bibitem{berant2013semantic}
Berant, J., Chou, A., Frostig, R., Liang, P.: Semantic parsing on freebase from
  question-answer pairs. In: EMNLP. pp. 1533--1544 (2013)

\bibitem{berant2014semantic}
Berant, J., Liang, P.: Semantic parsing via paraphrasing. In: Proceedings of
  ACL. vol.~7, p.~92 (2014)

\bibitem{bollacker2008freebase}
Bollacker, K., Evans, C., Paritosh, P., Sturge, T., Taylor, J.: Freebase: a
  collaboratively created graph database for structuring human knowledge. In:
  Proceedings of the 2008 ACM SIGMOD international conference on Management of
  data. pp. 1247--1250. ACM (2008)

\bibitem{cimiano2013multilingual}
Cimiano, P., Lopez, V., Unger, C., Cabrio, E., Ngomo, A.C.N., Walter, S.:
  Multilingual question answering over linked data (qald-3): Lab overview. In:
  Information Access Evaluation. Multilinguality, Multimodality, and
  Visualization, pp. 321--332. Springer (2013)

\bibitem{dima2013intui2}
Dima, C.: Intui2: A prototype system for question answering over linked data.
  Proceedings of the Question Answering over Linked Data lab (QALD-3) at CLEF
  (2013)

\bibitem{fader2013paraphrase}
Fader, A., Zettlemoyer, L.S., Etzioni, O.: Paraphrase-driven learning for open
  question answering. In: ACL (1). pp. 1608--1618. Citeseer (2013)

\bibitem{giannone2013hmm}
Giannone, C., Bellomaria, V., Basili, R.: A hmm-based approach to question
  answering against linked data. Proceedings of the Question Answering over
  Linked Data lab (QALD-3) at CLEF  (2013)

\bibitem{han2013umbc}
Han, L., Kashyap, A., Finin, T., Mayfield, J., Weese, J.: Umbc ebiquity-core:
  Semantic textual similarity systems. In: Proceedings of the Second Joint
  Conference on Lexical and Computational Semantics. vol.~1, pp. 44--52 (2013)

\bibitem{he2014question}
He, S., Liu, K., Zhang, Y., Xu, L., Zhao, J.: Question answering over linked
  data using first-order logic. In: Proceedings of Empirical Methods in Natural
  Language Processing (2014)

\bibitem{he2013casia}
He, S., Liu, S., Chen, Y., Zhou, G., Liu, K., Zhao, J.: Casia@ qald-3: A
  question answering system over linked data. Proceedings of the Question
  Answering over Linked Data lab (QALD-3) at CLEF  (2013)

\bibitem{joris2013scalewelis}
Joris, G., Ferr{\'e}, S.: Scalewelis: a scalable query-based faceted search
  system on top of sparql endpoints. In: Work. Multilingual Question Answering
  over Linked Data (QALD-3) (2013)

\bibitem{lehmann2014dbpedia}
Lehmann, J., Isele, R., Jakob, M., Jentzsch, A., Kontokostas, D., Mendes, P.N.,
  Hellmann, S., Morsey, M., van Kleef, P., Auer, S., et~al.: Dbpedia--a
  large-scale, multilingual knowledge base extracted from wikipedia. Semantic
  Web  (2014)

\bibitem{lopez2012poweraqua}
Lopez, V., Fern{\'a}ndez, M., Motta, E., Stieler, N.: Poweraqua: Supporting
  users in querying and exploring the semantic web. Semantic Web  3(3),
  249--265 (2012)

\bibitem{lopez2011question}
Lopez, V., Uren, V., Sabou, M., Motta, E.: Is question answering fit for the
  semantic web?: a survey. Semantic Web  2(2),  125--155 (2011)

\bibitem{manning2014stanford}
Manning, C.D., Surdeanu, M., Bauer, J., Finkel, J., Bethard, S.J., McClosky,
  D.: The stanford corenlp natural language processing toolkit. In: Proceedings
  of 52nd Annual Meeting of the Association for Computational Linguistics:
  System Demonstrations. pp. 55--60 (2014)

\bibitem{milne2013open}
Milne, D., Witten, I.H.: An open-source toolkit for mining wikipedia.
  Artificial Intelligence  194,  222--239 (2013)

\bibitem{pradel2013swip}
Pradel, C., Peyet, G., Haemmerl{\'e}, O., Hernandez, N.: Swip at qald-3:
  results, criticisms and lesson learned. Valencia, Spain  (2013)

\bibitem{shekarpour2013question}
Shekarpour, S., Ngonga~Ngomo, A.C., Auer, S.: Question answering on interlinked
  data. In: Proceedings of the 22nd international conference on World Wide Web.
  pp. 1145--1156. International World Wide Web Conferences Steering Committee
  (2013)

\bibitem{suchanek2007yago}
Suchanek, F.M., Kasneci, G., Weikum, G.: Yago: a core of semantic knowledge.
  In: Proceedings of the 16th international conference on World Wide Web. pp.
  697--706. ACM (2007)

\bibitem{unger2012template}
Unger, C., B{\"u}hmann, L., Lehmann, J., Ngonga~Ngomo, A.C., Gerber, D.,
  Cimiano, P.: Template-based question answering over rdf data. In: Proceedings
  of the 21st international conference on World Wide Web. pp. 639--648. ACM
  (2012)

\bibitem{xu2015longest}
Xu, K., Zhang, S., Feng, Y., Huang, S., Zhao, D.: What is the longest river in
  the usa? semantic parsing for aggregation questions. In: Twenty-Ninth AAAI
  Conference on Artificial Intelligence (2015)

\bibitem{yahya2012natural}
Yahya, M., Berberich, K., Elbassuoni, S., Ramanath, M., Tresp, V., Weikum, G.:
  Natural language questions for the web of data. In: Proceedings of the 2012
  Joint Conference on Empirical Methods in Natural Language Processing and
  Computational Natural Language Learning. pp. 379--390. Association for
  Computational Linguistics (2012)

\bibitem{yahya2013robust}
Yahya, M., Berberich, K., Elbassuoni, S., Weikum, G.: Robust question answering
  over the web of linked data. In: Proceedings of the 22nd ACM international
  conference on Conference on information \& knowledge management. pp.
  1107--1116. ACM (2013)

\bibitem{zou2014natural}
Zou, L., Huang, R., Wang, H., Yu, J.X., He, W., Zhao, D.: Natural language
  question answering over rdf: a graph data driven approach. In: Proceedings of
  the 2014 ACM SIGMOD international conference on Management of data. pp.
  313--324. ACM (2014)

\end{thebibliography}
\clearpage
\end{document}